\documentclass[conference]{IEEEtran}
\IEEEoverridecommandlockouts
\usepackage{amsmath,amssymb,amsfonts}
\usepackage{algorithmic}
\usepackage{graphicx}
\usepackage{textcomp}
\usepackage{xcolor}
\usepackage{times}  
\usepackage{helvet}  
\usepackage{courier}  
\usepackage[hyphens]{url}  
\usepackage{graphicx} 
\usepackage[square,sort,comma,numbers]{natbib}
\usepackage{caption} 
\DeclareCaptionStyle{ruled}{labelfont=normalfont,labelsep=colon,strut=off} 
\usepackage{tikz}
\usepackage{tablefootnote}

\newcommand*\circled[1]{\tikz[baseline=(char.base)]{
            \node[shape=circle,fill,inner sep=1pt] (char) {\footnotesize \textcolor{white}{#1}};}}

\usepackage{algorithmic}
\usepackage{algorithm}
\usepackage[algo2e,boxed,ruled]{algorithm2e} 
\SetKwRepeat{Do}{do}{while}
\usepackage{amsthm} 
\usepackage{mathtools}
\usepackage{booktabs}

\usepackage{longtable}
\usepackage[format=plain,indention=0cm, font=small, labelfont=bf]{caption}
\usepackage{tabu}

\usepackage{amssymb}
\usepackage{subfloat}

\usepackage[utf8]{inputenc} 
\usepackage[T1]{fontenc}    
\usepackage{hyperref}       
\usepackage{url}            
\usepackage{booktabs}       
\usepackage{amsfonts}       
\usepackage{nicefrac}       
\usepackage{microtype}      
\usepackage{xcolor}         
\usepackage{microtype}
\usepackage{amsmath}

\usepackage{times}
\usepackage{latexsym}

\usepackage{graphicx}
\usepackage{amsmath}
\usepackage{algorithmic}
\usepackage{algorithm}
\usepackage{subfig}
\usepackage{microtype}
\usepackage{color}

\usepackage{xurl}
\usepackage{multirow}
\usepackage{microtype}
\usepackage{color}
\usepackage{microtype}
\usepackage{subfloat}

\usepackage[utf8]{inputenc}

\usepackage{sidecap}
\usepackage{times}
\usepackage{latexsym}

\usepackage{amsmath}
\usepackage{algorithmic}
\usepackage{algorithm}
\usepackage{booktabs}

\usepackage{multirow}

\usepackage{newfloat}
\usepackage{listings}
\floatstyle{ruled}
\newfloat{listing}{tb}{lst}{}
\floatname{listing}{Listing}

\usepackage{bibentry}
\usepackage[draft]{todonotes}

\def\BibTeX{{\rm B\kern-.05em{\sc i\kern-.025em b}\kern-.08em
    T\kern-.1667em\lower.7ex\hbox{E}\kern-.125emX}}
\begin{document}

\title{Neurogenesis Dynamics-inspired Spiking Neural Network Training Acceleration}

\author{
    Shaoyi Huang  \textsuperscript{\rm 1},
    Haowen Fang,
    Kaleel Mahmood \textsuperscript{\rm 1},
    Bowen Lei \textsuperscript{\rm 2},
    Nuo Xu \textsuperscript{\rm 3},
    Bin Lei \textsuperscript{\rm 1},
    Yue Sun \textsuperscript{\rm 3}, \\
    Dongkuan Xu \textsuperscript{\rm 4}, 
    Wujie Wen \textsuperscript{\rm 3},
    Caiwen Ding \textsuperscript{\rm 1}\\
 $^1$University of Connecticut,
 $^2$Texas A\&M University,
$^3$Lehigh University,
 $^4$North Carolina State University\\
 \small \{shaoyi.huang, kaleel.mahmood, bin.lei, caiwen.ding\}@uconn.edu,\\
 \small bowenlei@stat.tamu.edu,
 \{nux219, yus516, wuw219\}@lehigh.edu,
 dxu27@ncsu.edu
 }

\maketitle

\begin{abstract} Biologically inspired  Spiking Neural Networks (SNNs) have attracted significant attention for their ability to provide extremely energy-efficient machine intelligence through event-driven operation  and sparse activities. 
As artificial intelligence (AI) becomes ever more democratized, there is an increasing need to execute SNN models on edge devices. 
Existing works adopt weight pruning to reduce SNN model size and accelerate inference. However, these methods mainly focus on how to obtain a sparse model for efficient inference, rather than training efficiency. 
To overcome these drawbacks, in this paper, we 
propose a 
\underline{N}eurogenesis
\underline{D}ynamics-inspired \underline{S}piking \underline{N}eural \underline{N}etwork training acceleration framework, NDSNN. Our framework is computational efficient and trains a model from scratch with dynamic sparsity without sacrificing model fidelity.
Specifically, we design a new drop-and-grow strategy with decreasing number of non-zero weights, to maintain extreme high sparsity and high accuracy.  We evaluate NDSNN using VGG-16 and ResNet-19 on CIFAR-10, CIFAR-100 and TinyImageNet. Experimental results show that NDSNN achieves up to 20.52\% improvement in accuracy on Tiny-ImageNet using ResNet-19 (with a sparsity of 99\%) as compared to other SOTA methods (e.g., Lottery Ticket Hypothesis (LTH), SET-SNN, RigL-SNN). In addition, the training cost of NDSNN 
is only 40.89\% of the LTH training cost on ResNet-19 and 31.35\% of the LTH training cost on VGG-16 on CIFAR-10.
\end{abstract}


\section{Introduction}






Biologically inspired  Spiking Neural Networks (SNNs) have attracted significant attention for their ability to provide extremely energy-efficient machine intelligence. SNNs achieve this performance through event-driven operation (e.g., computation is only performed on demand) and the sparse activities of spikes.
As artificial intelligence (AI) becomes ever more democratized, there is an increasing need to execute SNN models on edge devices with limited memory and restricted computational resources~\cite{finley2019democratization}. However, modern SNNs typically consist of at least millions to hundreds of millions of parameters (i.e., weights), which requires large memory storage and computations~\cite{fidjeland2010accelerated, qi2021accelerating, qi2021accommodating}. 
%
Therefore, it is desirable to investigate efficient implementation techniques for SNNs. 

\begin{figure}[!ht]
    \centering
    \includegraphics[width=1.0\columnwidth]{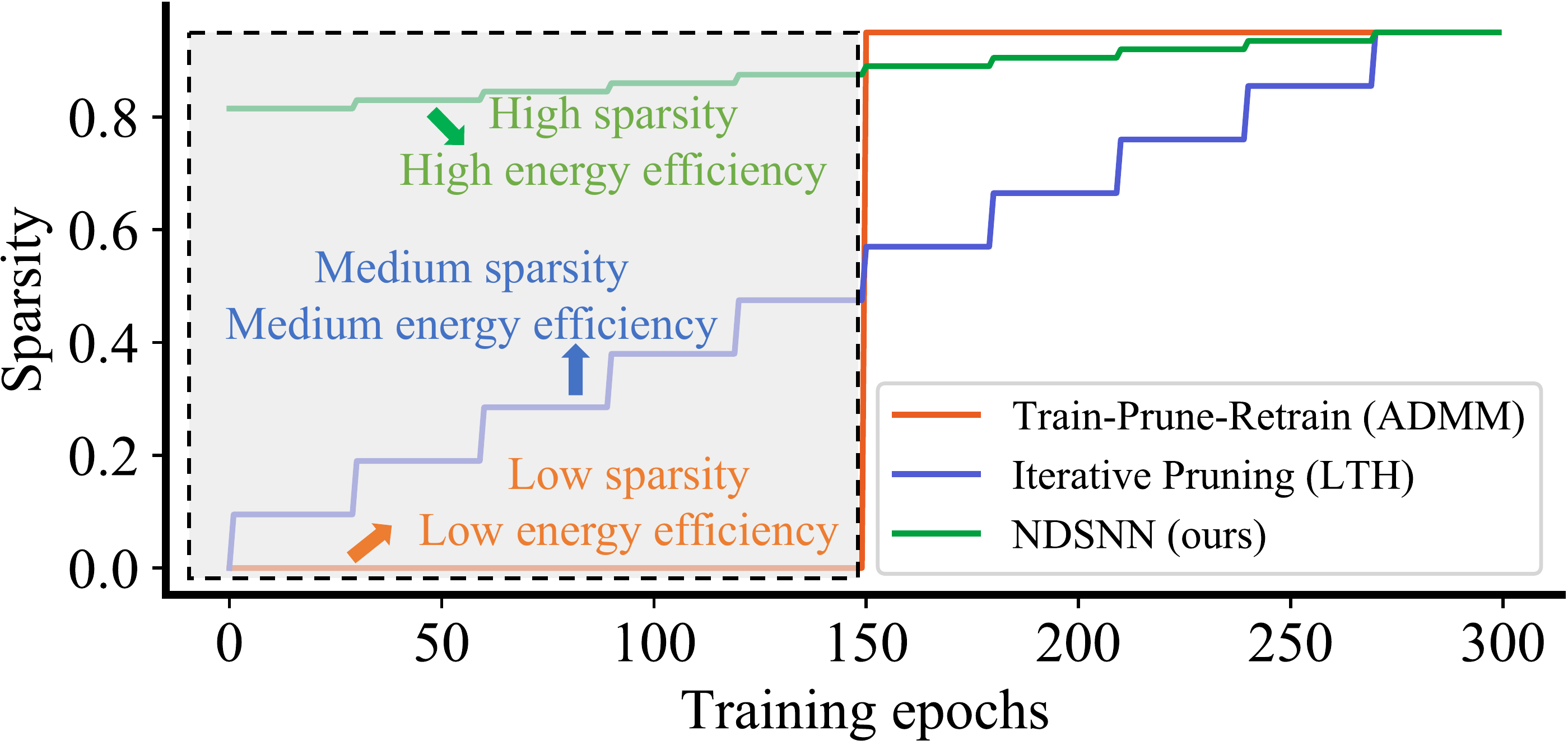}
    \caption{ Sparsity change of different sparsification methods on VGG-16 / ResNet-19 CIFAR-10. 
    }
    \label{fig:motivation}
\end{figure}

Recently, the use of sparsity to compress SNN model size and accelerate inference has attracted a surge of attention~\cite{deng2021comprehensive, kim2022exploring}, including the
train-prune-retrain method (e.g, alternating direction method of multipliers (ADMM) pruning~\cite{deng2021comprehensive, peng2021accelerating, chen2021re, luo2022codg}), iterative pruning (e.g., lottery ticket hypothesis (LTH)~\cite{frankle2018lottery,kim2022exploring})). The aforementioned methods are shown in Fig.~\ref{fig:motivation} and mainly focus on how to obtain a sparse model for efficient inference. However, the training process to obtain 
a sparse model is not efficient.
To illustrate consider the case VGG-16 on CIFAR-10, for train-prune-retrain~\cite{deng2021comprehensive, huang2022sparse} (orange line), the first 150 training epoches are dense (zero sparsity); For iterative pruning~\cite{kim2022exploring}, the sparsity gradually increases in the first 150 training epoches. As shown in the highlighted grey area, both methods have low sparsity hence low training efficiency.

In the field of neuroscience, the total number of neurons declines with age during the process of neuron's degeneration (i.e., old neuron's death) and redifferentiation (i.e., new neuron's birth), in human hippocampus, referred as Neurogenesis Dynamics~\cite{ming2011adult,spalding2013dynamics}. 
In this paper, inspired by the \underline{N}eurogenesis \underline{D}ynamics,
we 
propose an efficient \underline{S}piking \underline{N}eural \underline{N}etwork training acceleration framework, NDSNN. We analogize the neuron's death-and-birth renewal scheme to the drop-and-grow schedule in SNN sparse training. 
We dynamically reduce the number of neuron connections
in SNN sparse training, to reduce training memory footprint and improve training efficiency~\cite{davies2018loihi}.
The number of zeros decreases
in the dynamically changing process of weight mask tensor (i.e., a binary tensor which has the same size as weight, 0s / 1s denotes zeros / non-zeros in corresponding weight tensor). 
The sparsity during NDSNN training is illustrated in Fig.~\ref{fig:motivation} as the green curve.
We could train from a highly sparsified model (e.g., initial sparsity is 80\%) and achieve the final sparsity (e.g., 95\%).

\begin{figure*}[!ht]
    \centering
    \includegraphics[width=1.0\textwidth]{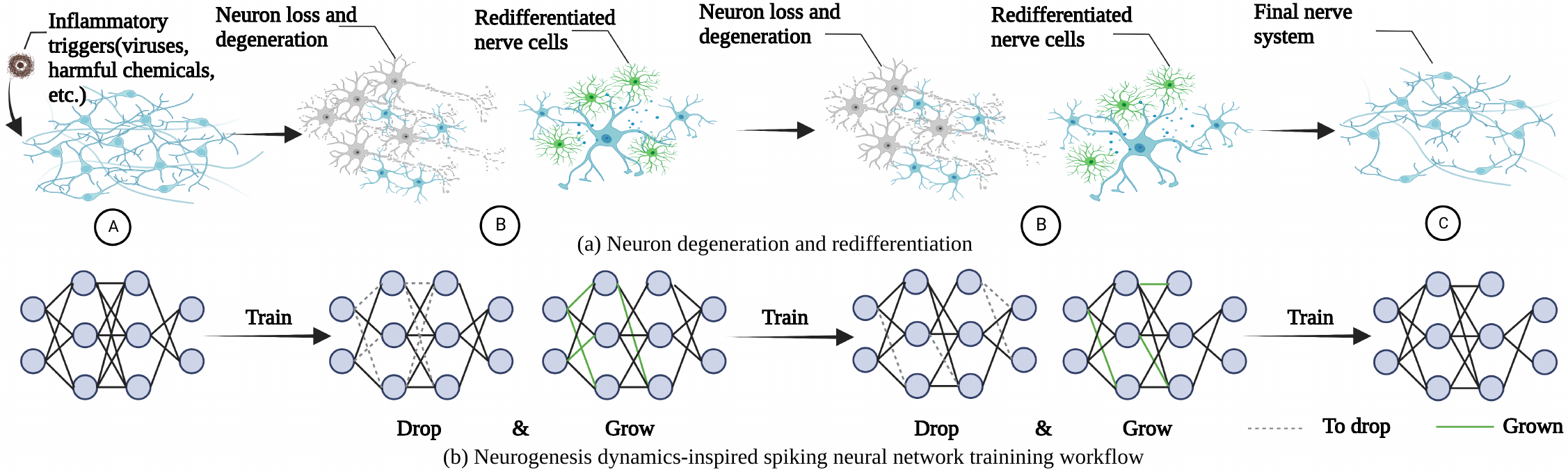}
    \caption{(a) shows the neurogenesis dynamics of nerve cells in the nervous system. \protect\circled{A}
    indicates inflammatory factors accumulating in nerve system. \protect\circled{B} indicates neuron degeneration and redifferentiation process. \protect\circled{C} is the final nerve system. (b) shows drop and grow process of the neural network. The total number of nonzero weights decreases with the increasing of drop-and-grow times.}
    \label{fig:workflow}
    \vspace{-0.5cm}
\end{figure*}



Overall our paper makes the following contributions:

\begin{itemize}

\item Inspired by neurogenesis dynamics, we propose an energy efficient spiking neural network training workflow.

\item To reach high sparsity and high energy efficiency with dense model like accuracy, we design a new drop-and-grow strategy with decreasing number of non-zero weights in the process of dynamically updating sparse mask.

\item We evaluate the training efficiency of NDSNN via normalizing spike rate. Results show that the cost of NDSNN on ResNet-19 and VGG-16 is 40.89\% and 31.35\% of state-of-the-art (SOTA), respectively.

\item We demonstrate extremely high sparsity (i.e., 99\%) model performance in SNN based vision tasks with acceptable accuracy degradation.

\end{itemize}

 We evaluate NDSNN using VGG-16 and ResNet on CIFAR-10, CIFAR-100 and TinyImageNet. Experimental results show that NDSNN achieves even high accuracy than dense model for ResNet-19 on CIFAR-10. On Tiny-ImageNet, NDSNN achieves up to 20.52\% increase in accuracy compared to the SOTA at a sparsity of 99\%. The training cost of NDSNN VGG-16 is 10.5\% of training a dense model.

\section{Related Work and Background}
\subsection{Related Work on Sparsity Exploration in SNN}

Several network compression schemes for SNNs have been proposed. In~\cite{deng2021comprehensive} the alternating direction method of multipliers (ADMMs) pruning is employed to compress the SNNs on various datasets. However, this technique has significant accuracy loss, especially when the model has high sparsity. 
Although IMP could find highly sparse neural network with high accuracy, it is time consuming (e.g. it takes 2720 epochs to achieve 89.91\% sparsity on both CIFAR-10 and CIFAT-100)~\cite{kim2022exploring}.
In~\cite{rathi2018stdp} they propose a Spike Timing Dependent Plasticity (STDP) based pruning method. Connections between pre-synaptic and post-synaptic neurons with low spike correlation are pruned. The correlation is tracked by STDP algorithm. The performance of this method is limited as the original model only achieves 93.2\% accuracy on MNIST, and accuracy drops to 91.5\% after 92\% weights are pruned.
In~\cite{nguyen2021connection} they propose a technique to prune connections during training. Weights will be pruned if they are less than a certain threshold or decrease significantly in a number of training iterations. However, the method's evaluation is limited, as it is only tested on a single dataset Caltech-101.


%





\vspace{-0.15cm}

\subsection{Spiking Neural Network}

A key difference of SNN from DNN is that spiking neuron is a stateful system that can be modeled by different equations. The commonly used Leaky Integrate and Fire (LIF) spiking neuron is defined as follows. 

\vspace{-0.25cm}
\begin{subequations}
\label{eq:neuron_model}
\begin{gather}
        v[t] = \alpha v[t{-}1] + \sum_i w_i s_i[t]  - \vartheta o[t-1] \label{eq:voltage} \\
        o[t]  = u(v[t] - \vartheta)  \label{eq:threshold} \\
        u(x) = 0, x < 0 \text{ otherwise 1} \label{eq:heaviside}
\end{gather}
\end{subequations}
\vspace{-0.5cm}

where $t$ indicates time. Eq.~\eqref{eq:voltage} depicts the dynamics of the neuron's membrane potential $v[t]$. $\alpha \in (0,1]$ determines $v[t]$ the decay speed. $s_i[t] \in \{0,1\}$ is a sequence which consists of only 0 and 1 to represent the $i-{th}$ input spike train and $w_i$ is the corresponding weight. $o[t] \in \{0,1\}$ is the neuron's output spike train, $u(x)$ is the Heaviside step function.




\begin{figure*}[!ht]
    \centering
    \includegraphics[width=1.0\textwidth]{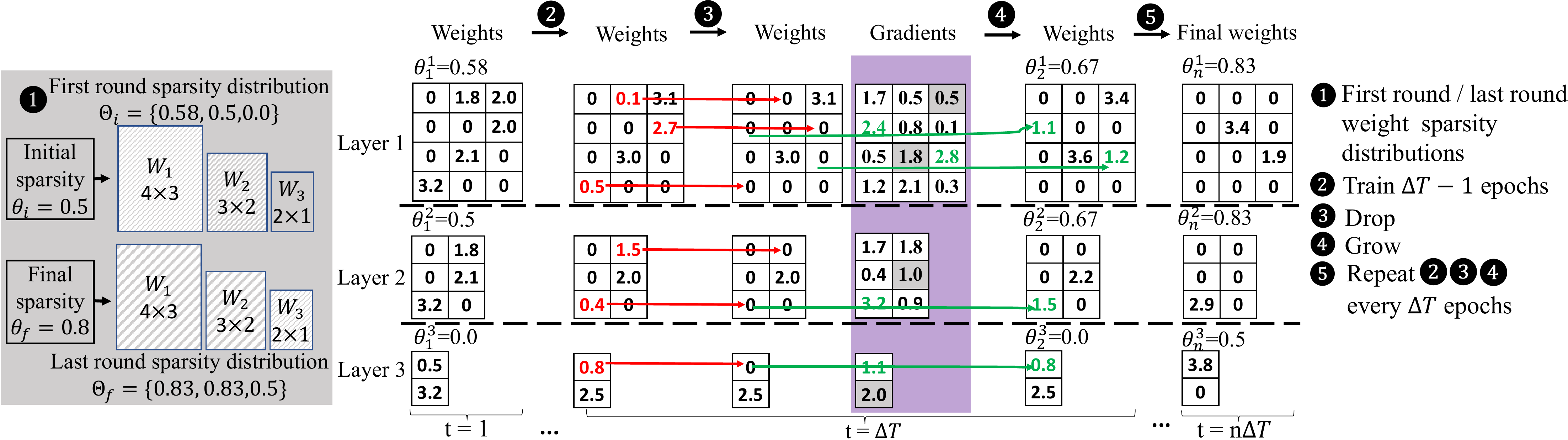}
    \caption{A toy example of NDSNN training process. Red arrows denote dropping weights and green arrows denote growing weights. }
    \label{fig:process}
\vspace{-0.5cm}
\end{figure*}

Note that Eq.~\eqref{eq:voltage} is recursive in the temporal domain, so it is possible to use Backpropagation Through Time (BPTT) to train SNNs. However, an issue arises with Eq.~\eqref{eq:heaviside}, whose derivative is the Dirac Delta function $\Delta(x)$. To overcome this, surrogate gradient method can be used~\cite{neftci2019surrogate} so that the derivative of $u(x)$ is approximated by the derivative of a smooth function. In the forward pass, the SNN still outputs spikes, while in backward pass, $\Delta(x)$ is replaced by a surrogate function so the Heaviside step function has an approximate derivative.

The BPTT for SNNs using a surrogate gradient is derived as follows. Let $L$ be the loss, $\delta_l[t] {=} \frac{\partial L}{\partial o_l[t]}$ be the error signal at layer $l$ time step $t$, $\epsilon_l[t] {=} \frac{\partial L}{\partial v_l[t]}$. $\delta_l[t]$ is propagated recursively as following rules, and gradient of $l^{th}$ layer weight $w_l$ is calculated using Eq.~\eqref{eq:gradient}.

\vspace{-0.2cm}
\begin{subequations}
\label{eq:BPTT}
\begin{gather}
    \delta_l[t] = \epsilon_{l+1}[t] w_{l+1} \\
    \epsilon_{l}[t] = \delta_{l}[t] \phi_{l}[t] + \alpha \epsilon_{l}[t] \\
    \frac{\partial L}{\partial w_l} = \sum_{t=0}^{T-1} \epsilon_l[t] \cdot [s_l[t]]^\intercal \label{rq:grad}
    \end{gather}
\end{subequations}
where $\phi_l[t] {=} \frac{\partial o_l[t]}{\partial v_l[t]} {=} \frac{\partial u(v_l[t] - \vartheta)}{\partial v_l[t]}$. Note that $u(x)$ does not have a well-defined derivative, so we use the gradient surrogate function proposed in~\cite{fang2021deep} to approximate it, such that:

\begin{equation}
    \frac{\partial u(x)}{\partial x} \approx \frac{1}{1 + \pi^2 x^2}
\label{eq:gradient} 
\end{equation}


\section{Neurogenesis Dynamics-inspired Sparse Training on SNN}

We illustrate the overall workflow of the biological and corresponding computational methods in Fig. ~\ref{fig:workflow}.


\subsection{Analogizing Neurogenesis Dynamics in Sparse Training}

In human hippocampus, the total number of neurons declines with age during the process of neuron's degeneration (i.e., old neuron's death) and redifferentiation (i.e., neuron's birth)~\cite{spalding2013dynamics}.
We analogize the neuron's death-and-birth renewal scheme to the drop-and-grow schedule in sparse training~\cite{liu2021we, liu2021sparse, peng2022towards, huang2022dynamic}. Here \textit{drop} means the insignificant connections are deactivated (weights with least absolute magnitude are set as zeros). In our formulation \textit{grow} refers to creating new connections (weights with high importance are updated to nonzeros). For the dynamics of neurogenesis in the human hippocampus, the neurons declines with age~\cite{spalding2013dynamics}. Similarily in our framework, we reduce the number of connections or reduce the number of activated weights in the sparse training process in consideration of the memory limitation of neuromorphic chips~\cite{davies2018loihi}. 

\subsection{Problem Definiton}

We aim to achieve
high sparsity (low memory overhead) during training and 
high energy efficiency (through SNN implementation) without noticeable accuracy loss.
The problem is formally defined as: consider a $L$-layer SNN with dense weights $W = [W_1, W_2, ... , W_L]$, a dataset $\mathcal{X}$, and a target sparsity $\theta_f$, our goal is to develop an training workflow such that the training process requires less memory overhead and less computation, and the trained model achieves high accuracy.



\subsection{Neurogenesis Dynamics-inspired Spiking Neural Network (NDSNN) Training Acceleration Framework}

Fig. ~\ref{fig:workflow} shows the overview of neurogenesis dynamics-inspired spiking neural network (NDSNN) workflow. Fig.~\ref{fig:workflow}(a) demonstrates the neuron cell loss or degeneration (the grey neuron cells) and redifferentiation process (the green neuron cells). In Fig.~\ref{fig:workflow}(b) we illustrate the training process of NDSNN, where we drop the weights (i.e., setting the smallest positive weights and the largest negative weights as zeros) in grey color and grow the weights (i.e., update the zeros weights to nonzeros) in green color, every $\Delta T$ iterations. The number of weights we dropped is larger than the grown ones each drop-and-grow schedule. Thus, the number of nonzero weights decreases with the increasing of drop-and-grow times.

The goal of the proposed training method is to reduce memory footprint and computations during the whole training process. To achieve it, our proposed method uses less weights and gradients than SOTA methods via dynamically updating the sparse mask and training from scratch. Specifically, we denote $\theta_i$ and $\theta_f$ as the initial and target sparsity, respectively. $t_0$ is the starting step of training, $\Delta T$ is the pruning frequency. The full training workflow is be formed in the following steps.


\circled{1} \textbf{First round / last round weight sparsity distributions across different layers.} Let $\Theta_i = {\theta_i^1, \theta_i^2, ..., \theta_i^L}$ denote the initial sparsity distribution (i.e., sparsity of different layers at the beginning of training) of SNN model and $\Theta_f = {\theta_f^1, \theta_f^2, ..., \theta_f^L}$ denote the final sparsity distribution (i.e., sparsity of different layers at the end of training) of the model. Here, we use ERK~\cite{mocanu2018scalable} to distributing the non-zero weights across the layers while maintaining the overall sparsity. We denote $n^l$ as the number of neurons at $l$-th layer and $w^l$, $h^l$ as the width and height of the $l$-th convolutional kernel,then the number of parameters of the sparse convolutional layers are scaled proportional to $1-\frac{n^{l-1}+n^l+w^l+h^l}{n^{l-1}*n^l*w^l*h^l}$. In our case, the overall sparsity at the beginning of training $\theta_i$ is less than the one at the end of training $\theta_f$. Following the same scaling proportion distribution, the sparsity of each separate convolutional layer at the beginning of training is smaller than it's sparsity at the end of training (i.e., for $l$-th layer, we have $\theta_i^l \le \theta_f^l$). The sparsity of $l$-th layer at $t$-th iteration is formulated as: 
\begin{equation}
\begin{aligned}
\theta_t^l &= \theta_f^l + (\theta_i^l - \theta_f^l)(1 - \frac{t - t_0}{n\Delta t})^3,\\ t\in \{ t_0, t_0&+\Delta T, ..., t_0+n\Delta T\}, l\in \{1, 2, ..., L \}.
\label{eq:gradual} 
\end{aligned}
\end{equation}


\circled{2} \textbf{Training.} We define non-active weights as weights has value of zeros and active weights as weights has value of non-zeros.
For each iteration, we only update the active weights. 
In backward path, gradients are calculated using BPTT with surrogate gradient method, and forward pass is carried out like standard neural network training. 

\circled{3} \textbf{Dropping (neuron death).} During training, the sparse masks are updated every $\Delta T$ iteration, i.e., for $l$-th layer, we drop $D_d^l$ 
weights that are closest to zero (i.e., the smallest positive weights and the largest negative weights). 
we denote $d_0$ as the initial death ratio (i.e., the ratio of weights to prune from non-zeros) and $d_t$ as the death ratio at step $t$. We use the cosine annealing learning rate scheduler~\cite{loshchilovsgdr} for death ratio updating. Then, we have
\begin{equation}
\small
\begin{aligned}
d_t = &{d}_{min} + 0.5(d_0 - {d}_{min})(1+cos(\frac{\pi t}{n\Delta t})), \\
t&\in \{ t_0, t_0+\Delta T, ..., t_0+n\Delta T\},
\label{eq:death_rate_decay} 
\end{aligned}
\end{equation}

where ${d}_{min}$ is the minimum death rate during the training. At $q^{th}$ round, the number of 1s in sparse mask of $l$-th layer ${N_{pre}}_q^l$ before dropping is

\begin{equation}
\small
\begin{aligned}
{N_{pre}}_q^l = N^l(1 - \theta_{q-1}^l), 1 \leq q \leq n, l\in \{1, 2, ..., L \}
\label{eq:pre} 
\end{aligned}
\end{equation}
where $N^l$ is the number of all weight elements in $l$-th layer and $\theta_{q-1}^l$ is the training sparsity of $l$-th layer at $(q-1)$-th round. We denote the number of dropped weights of $l$-th layer at $q$-th round as $D_q^l$, then, we have 
\begin{equation}
\small
\begin{aligned}
D_q^l = d_t \times {N_{pre}}_q^l, 1 \leq q \leq n,l\in \{1, 2, ..., L \}.
\label{eq:num_drop} 
\end{aligned}
\end{equation}


\circled{4} \textbf{Growing (neuron birth).} After dropping weights, the number of 1s in $l$-th layer sparse mask ${N_{post}}_q^l$ is

\begin{equation}
\small
\begin{aligned}
{N_{post}}_q^l = {N_{pre}}_q^l - D_q^l, 1 \leq q \leq n, l\in \{1, 2, ..., L \}.
\label{eq:post} 
\end{aligned}
\end{equation}

Combining Equation ~\ref{eq:gradual} and ~\ref{eq:post}, we obtain the number of weights to be grown, which is denoted as $G_q^l$, we have

\begin{equation}
\small
\begin{aligned}
G_q^l = N^l - {N_{post}}_q^l - \theta_t^l \times N^l, 
1 \leq q \leq n, l\in \{1, 2, ..., L \}.
\label{eq:num_grow} 
\end{aligned}
\end{equation}

The toy example of the training process is shown in Fig.~\ref{fig:process}.

\begin{algorithm}[!ht]
\scriptsize
  \caption{NDSNN training flow.}
  \label{alg:dst-snn}
\begin{algorithmic}
    \STATE {\bfseries Input:} a $L$-layer SNN model $g$ with dense weight $\textbf{W} = {\textbf{W}_1, \textbf{W}_2,  ..., \textbf{W}_L}$, input data $\mathcal{X}$, update frequency $\Delta T$, initial sparsity $\theta_i$, final sparsity $\theta_f$, learning rate $\alpha$, total number of training iterations $T_{end}$.
    
    
    \STATE {\bfseries Set} ${\textbf{M}_1, \textbf{M}_2,  ..., \textbf{M}_L}$ as the sparse masks. 
    
    \STATE {\bfseries Output:} a $L$-layer sparse network with sparsity distribution $P_f$.

  \STATE Calculate $\Theta_i = {\theta_i^1, \theta_i^2, ..., \theta_i^L}$ and $\Theta_f = {\theta_f^1, \theta_f^2, ..., \theta_f^L}$ using initial sparsity $\theta_i$ and final sparsity $\theta_f$, respectively via ERK.
  
  \STATE $\textbf{W}'={\textbf{W}'_1, \textbf{W}'_2,  ..., \textbf{W}'_L} \leftarrow$ sparsify ${\textbf{W}_1, \textbf{W}_2,  ..., \textbf{W}_L}$ with $P_i$
    \FOR{each training iteration $t$ }
        \STATE Loss $E$ $\leftarrow$ $g(x_t,\textbf{W}')$, $x_t \in \mathcal{X}$
        \IF {$t$ (mod $\Delta T$) == 0 and $t$ $<$ $T_{end}$}
            \FOR{$1 \leq l \leq L$}
                \STATE Calculate the number of weights to drop $D_{t/\Delta T}^l$ using Equation~\ref{eq:death_rate_decay}~\ref{eq:pre}~\ref{eq:num_drop}
                \STATE $\textbf{W}'_i \leftarrow$ ArgDrop$(\textbf{W}'_i, $ArgTopK$(\textbf{W}'_i, D_{t/\Delta T}^l))$
                \STATE Calculate the number of weights to grow $G_{t/\Delta T}^l$ using Equation~\ref{eq:post}~\ref{eq:num_grow}
                \STATE Calculate gradient $\textbf{Grad}_l$ by equation (~\ref{rq:grad})
                \STATE  $\textbf{W}'_l \leftarrow $ArgGrow$(\textbf{W}'_l, $ArgTopK$(\textbf{Grad}_l \cdot (\textbf{M}_l==0), G_{t/\Delta T}^l))$
            \ENDFOR
        \ELSE
            \STATE $\textbf{W}'_l \leftarrow \textbf{W}'_l - \alpha \nabla(\textbf{W}'_l) \delta_t$
        \ENDIF
  \ENDFOR
\end{algorithmic}
\end{algorithm}

\begin{table*}[!ht]
\small
\centering
\resizebox{0.9\textwidth}{!}{
\begin{tabular}{|l|cccc|cccc|cccc|}
\toprule
\textbf{Dataset} && \textbf{CIFAR-10} &&&&\textbf{CIFAR-100} &&&&\textbf{Tiny-ImageNet} && \\
\midrule
\textbf{Sparsity ratio} & 90\% & 95\% & 98\% & 99\% & 90\% & 95\% & 98\% & 99\% & 90\% & 95\% & 98\% & 99\%  \\
\midrule
\textbf{VGG-16(Dense)} &  &92.59  & & &  &69.86 & & &  &39.45 & &  \\
\midrule
LTH-SNN~\cite{frankle2018lottery} & 89.77 & 89.97 & 88.97 & 88.07 & 64.41 & 64.84& 62.97 & 51.31 & 38.01 &37.51 &35.66 &30.98   \\
\midrule
SET-SNN~\cite{mocanu2018scalable} & 91.22 & 90.41 & 87.26 & 83.40 & 66.52 & 63.48& 58.04 & 50.83 & 38.80 &37.34 & 33.40& 26.74  \\
RigL-SNN~\cite{evci2020rigging} & 91.64 & 90.06 & 87.30 & 84.08 & 66.59 & 63.47& 58.21 & 52.26 & 38.96 & 37.75& 32.94& 28.39  \\
NDSNN (Ours) & \textbf{91.84} & \textbf{91.31} & \textbf{89.62} & \textbf{88.13} & \textbf{68.07} & \textbf{66.73} & \textbf{63.51} & \textbf{58.07} & \textbf{39.12} &\textbf{37.77} & \textbf{36.23}& \textbf{33.84}  \\
\midrule
\textbf{ResNet-19(Dense)} &  &  91.10& & &  & 71.94& &  &  &50.32 & &  \\
\midrule
LTH-SNN~\cite{frankle2018lottery} & 87.57 & 87.16 & 85.91 & 82.29 & 54.66 & 54.78& 42.10 & 41.46 & 38.40 &37.74 &31.34 &21.44 \\
\midrule
SET-SNN~\cite{mocanu2018scalable} & 90.79 & 90.07 & 87.24 & 83.17 & 68.12&64.65&57.49& 49.11 & 49.46 & 42.13& 37.25& 27.79\\
RigL-SNN~\cite{evci2020rigging} & 90.69 & 90.02& 87.19 & 83.26 & 67.33& 65.23 & 56.96 & 47.96 & \textbf{49.49} &40.40 & 37.98& 24.13\\
NDSNN (Ours) & \textbf{91.13} & \textbf{90.47} & \textbf{88.61} & \textbf{86.30} & \textbf{70.08} & \textbf{68.95} & \textbf{65.48} & \textbf{59.61} & 49.25 & \textbf{47.45}& \textbf{45.09} & \textbf{41.96} \\
\bottomrule
\end{tabular}
}
\caption{Test accuracy of sparse VGG-16 and ResNet-19 on CIFAR-10, CIFAR-100, Tiny-ImageNet datasets. The highest test accuracy scores are marked in bold.
 The LTH-SNN results are our reproduced accuracy using method from ~\cite{kim2022exploring}. 
 }
\label{tb:cifar}
\vspace{-0.5cm}
\end{table*}

\subsection{Memory Footprint Analysis}


We further investigate the training efficiency of our proposed method in terms of memory footprint.
Suppose a sparse SNN model
with a sparsity ratio (the percentage of number of zeros in weight) of $\theta \in [0, 1]$.
In each round of forward and backward propagation, $N$ weights and $tN$ gradients are saved.
For training, we use single precision (FP32) for weights and gradients to guarantee training accuracy. For inference, the weight precision $b_w$ is platform/implementation specific, for example Intel Loihi uses 8 bits~\cite{davies2018loihi}, mixed-signal design HICANN~\cite{schemmel2008wafer} has 4 bits for weights, FPGA-based designs such as \cite{panchapakesan2022syncnn} employes mixed precision (4 bits - 16 bits). For sparse models, we use indices (denoted by $b_{idx}$-bit numbers) to represent the sparse topology of weights/gradients within the dense model.
Compressed sparse row (CSR) is a commonly used sparse matrix storage format. 

Consider a 2-D weight tensor reshaping from a 4-D tensor. Each row of the 2-D weight tensor denotes the weight from a filter. For the $l$-th layer, we denote $F_l$, $Ch_l$, and $K_l$ as the number of filters (output channels), number of channels (input channels), and kernel size, respectively. Thus, the size of the weight matrix is $F_l$ rows by $Ch_l \cdot K_l^2$ columns.
Thus, the total number of indices of the entire network is $(1 - \theta) \cdot N + \sum_l (F_l + 1)$.  And the memory footprint of model representation together with gradients for unstructured sparsity is $(1-\theta)\cdot ((1+t)N \cdot b_w + N\cdot b_{idx})+\sum_l ((F_l + 1)\cdot b_{idx})$. Since the number of filters is much smaller than the total number of weights, we approximate the memory footprint as $(1-\theta)\cdot ((1+t)N \cdot b_w + N\cdot b_{idx})$.
Given same timestep $t$, higher sparsity means the lower memory overhead, which support the effectiveness of proposed method in reducing training memory since it has much higher training sparsity than SOTAs.



\section{Experimental Results}


\subsection{Experimental Setup}
\subsubsection{Architectures and Datasets.} We evaluate NDSNN on two popular neural network architectures (i.e., VGG-16 and ResNet-19) for three datasets (i.e., CIFAR-10, CIFAR-100 and Tiny-ImageNet). For fair comparison, we set the total number of training epochs as 300 on both CIFAR-10 and CIFAR-100, while as 100 on Tiny-ImageNet as LTH-SNN. We use SGD as the optimizer while setting the momentum as 0.9 and weight decay as $5e-4$. Also, we follow the setting in~\cite{kim2022exploring} and set the training batch size as 128, initial learning rate as $3e-1$ and timesteps as 5 across all experiments. 

\subsubsection{Baselines.} 
We train VGG-16 / ResNet-19 dense SNNs on various datasets and use them as our dense baselines. 
Other baselines are divided into two types based on the initial sparsity status of the training process (i.e., dense or sparse). For the former, we choose the SOTA pruning methods (i.e., LTH and ADMM) on SNN. For the latter, we implement the 
sparse training methods (i.e., SET~\cite{mocanu2018scalable}, RigL~\cite{evci2020rigging}) on SNN models (i.e., SET-SNN, RigL-SNN).

\subsubsection{Evaluation Platform} We conduct all experiments using PyTorch with CUDA 11.4 on Quadro RTX6000 GPU and Intel(R) Xeon(R) Gold 6244 @ 3.60GHz CPU. We use SpikingJelly~\cite{SpikingJelly} package for SNNs implementation.

\subsection{Accuracy Evaluations of NDSNN} 

\begin{figure*}[!ht]
    \centering
    \includegraphics[width=\textwidth]{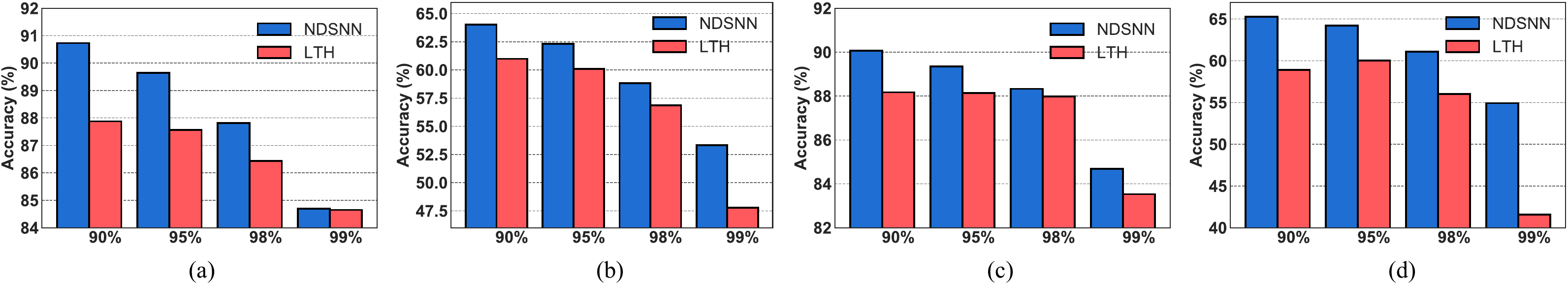}
    \caption{ Comparison of the accuracy of NDSNN and LTH for different sparsity when trained with smaller timestep (timestep=2) on different models and datasets. (a) VGG-16/CIFAR-10. (b) VGG-16/CIFAR-100. (c) ResNet-19/CIFAR-10. (d) ResNet-19/CIFAR-100}
    \label{fig:smaller_timestep}
\vspace{-0.5cm}
\end{figure*}

\subsubsection{CIFAR-10 and CIFAR-100} Evaluation results on CIFAR-10 and CIFAR-100 using VGG-16 and ResNet-19 are shown in Table~\ref{tb:cifar}. We compare NDSNN with baselines at sparsity ratios of 90\%, 95\%, 98\% and 99\% on different models and datasets. 
Experimental results show that NDSNN outperforms the SOTA baselines on each dataset for VGG-16 and ResNet-19.
    Specifically, on CIFAR-100, for VGG-16, our proposed method has up to 3.66\%, 3.26\%, 5.47\%, 7.24\% increase in accuracy (that is relatively 5.68\%, 5.14\%, 9.42\% and 14.24\% higher accuracy) at four different sparsity, respectively. While for ResNet-19, NDSNN has 15.42\%, 14.17\%, 23.88\% and 18.15\% increase in accuracy (that is relatively 28.2\%, 14.17\%, 23.38\%, 18.15\% higher accuracy) compared to LTH-SNN, obtains 1.96\%, 4.30\%, 7.99\%, 10.5\% higher accuracy than SET-SNN and achieves 2.75\%, 3.72\%, 8.52\%, 11.65\% higher accuracy than RigL-SNN at a sparsity of 90\%, 95\%, 98\% and 99\%, respectively. On CIFAR-10, for VGG-16, NDSNN has up to 2.07\%, 1.34\%, 2.36\%, 4.73\% relatively higher accuracy than SOTA at sparsity of 90\%, 95\%, 98\% and 99\%, respectively. While for ResNet-19, NDSNN has even higher accuracy than the dense model at a sparsity of 90\% and achieves the highest accuracy compared to other baselines at different sparsity.


\subsubsection{Tiny-ImageNet} The accuracy results on Tiny-ImageNet are shown in Table~\ref{tb:cifar}. Overall, for both VGG-16 and ResNet-19, NDSNN outperforms
other baselines. More specifically, for VGG-16, NDSNN has up to 7.1\% higher accuracy than other methods at a sparsity of 99\%. For ResNet-19, NDSNN has 10.85\%, 9.71\%, 13.75\%, 20.52\% higher accuracy than LTH-SNN at sparsity of 90\%, 95\% and 98\%, 99\%, respectively. Compared to SET-SNN, NDSNN has 7.10\% and 14.17\% increase in accuracy at the sparsity of 99\% for VGG-16 and ResNet-19, independently. 
Compared to RigL-SNN, NDSNN has up to 5.45\% and 17.83\% increase in accuracy at a sparsity of 99\% for VGG-16 and ResNet-19, respectively.

\subsubsection{Comparison with ADMM Pruning} We compare NDSNN with ADMM pruning using data from ~\cite{deng2021comprehensive} as shown in Table~\ref{tb:ADMM}. It can be seen that the accuracy loss become noticeable when the sparsity reaches 75\% on CIFAR-10 using LeNet-5. However, the accuracy loss is almost 0 on CIFAR-10 using VGG-16 at the sparsity of 75\% which indicates that NDSNN has less accuracy loss when achieving the same sparsity.

\begin{table}[!ht]
\small
\centering
\resizebox{0.8\columnwidth}{!}{
\begin{tabular}{|l|cccc|}
\toprule
\textbf{Dataset} & &\textbf{CIFAR-10} && \\
\midrule
\textbf{Sparsity ratio} &40\% & 50\% & 60\% & 75\%  \\
\midrule
\textbf{LeNet-5(Dense)} &  &89.53 & & \\
\midrule
ADMM~\cite{deng2021comprehensive} & 89.75 & 89.15 & 88.35 & 87.38 \\
\midrule
Acc. Loss (\%) & 0.18 & -0.38 & -1.18 & -2.15 \\
\midrule

\textbf{VGG-16(Dense)} &  &92.59 & & \\
\midrule
NDSNN (ours) & 92.46 & 92.32 & 92.33 & 92.18 \\
\midrule
Acc. Loss (\%) & -0.001 & -0.003 & -0.003 & -0.004 \\
\bottomrule
\end{tabular}
}
\caption{Comparison of ADMM with NDSNN on CIFAR-10.}
\label{tb:ADMM}
\end{table}

\subsection{Efficiency Evaluations of NDSNN} 

\begin{figure}[!ht]
    \centering
    \includegraphics[width=\columnwidth]{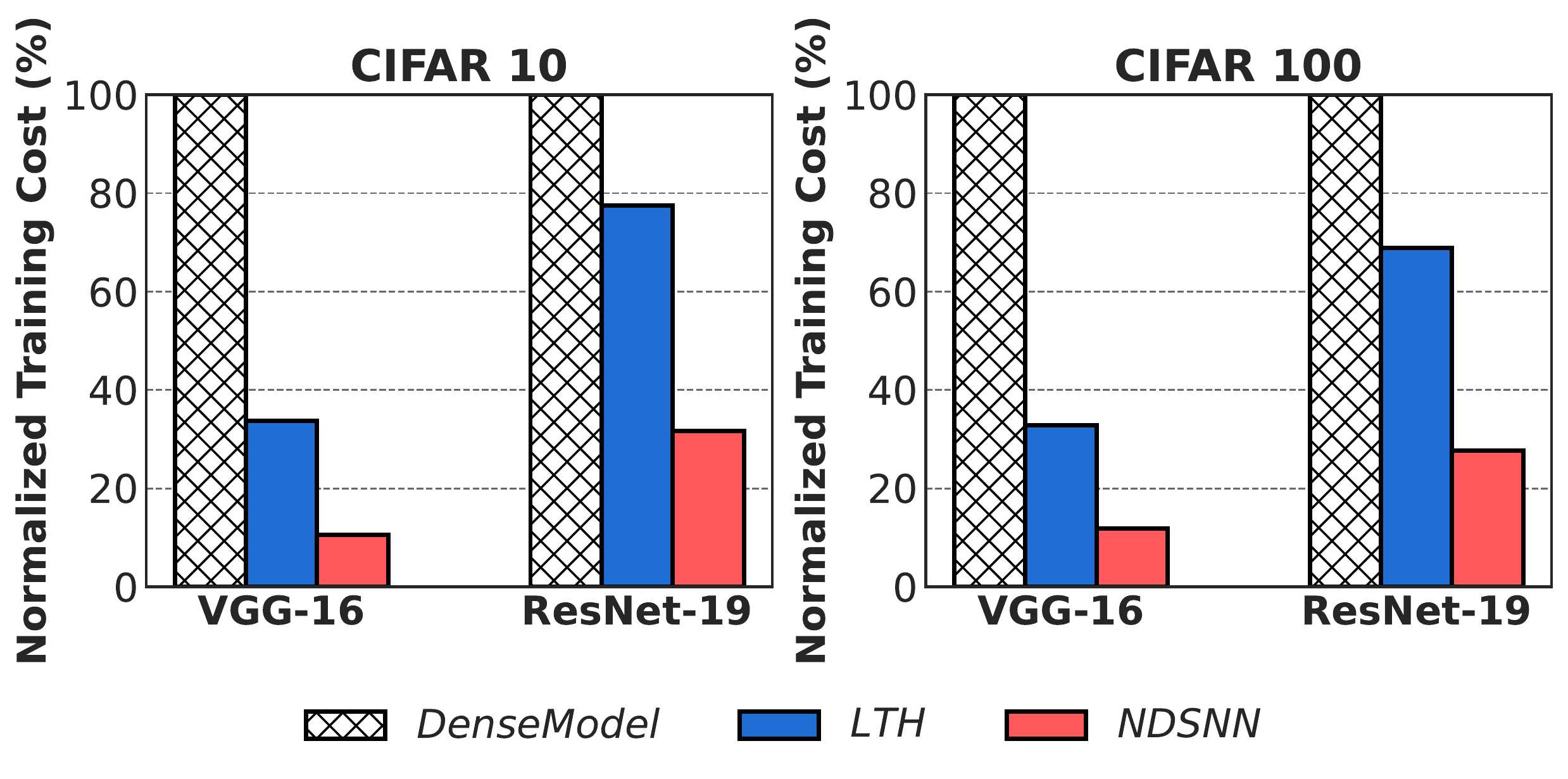}
    \caption{ Training cost comparison on CIFAR-10/CIFAR-100 using VGG-16 and ResNet-19.
    }
    \label{fig:trainingcost}
\end{figure}

We 
quantitatively analyze the training cost of dense SNN model, LTH and NDSNN, as showed in Fig.~\ref{fig:trainingcost}. Since no computation is required if there is no input spikes or a connection is pruned. Such that the relative computation cost of sparse model with respect to dense model at training epoch $i$ can be calculated as:
   $[
    {R^i_{s} \times Sparsity_i}]/{R^i_{d}}$,
where $R^i_{s}$ or $R^i_{d}$ is the average spike rate of the sparse model (LTH/NDSNN) or the dense model at epoch $i$, which can be tracked throughout entire training. $Sparsity_i$ is the sparsity of the model. On CIFAR 10, the training cost of NDSNN VGG-16 is 10.5\% of training a dense model. The cost of NDSNN on ResNet-19 and VGG-16 is 40.89\% and 31.35\% of LTH, respectively. On CIFAR 100, the training cost of NDSNN  ResNet-19 is 27.63\% and 40.12\% of dense model and LTH respectively; The training cost of NDSNN VGG-16 is 11.87\% and 36.16\% of dense model and LTH respctively.

\subsection{Design Exploration}

\subsubsection{Effects of Different Initial Sparsity} As the initial sparsity has influence on the average training sparsity, thus the overall training cost. we study the effects of different initial sparsity on accuracy and training FLOPs. Experimental results on VGG-16 / ResNet-19 models and CIFAR-10 / CIFAR-100 datasets are shown in Table~\ref{tb:diff_init_sparsity}. It's observed that the accuracy gap is small for different initial sparsity. For high training sparsity, we choose initial sparsity from \{0.6, 0.7, 0.8\} for experiments on CIFAR-10 / CIFAR-100 / TinyImageNet.

\begin{table}[!ht]
\footnotesize
\addtolength{\tabcolsep}{-2pt}
\renewcommand{\arraystretch}{1}
\centering
\vspace{-0cm}
\resizebox{0.9\columnwidth}{!}{\begin{tabular}{|l|c|c|c|c|c|c}
\toprule
Target  & Initial  & VGG-16 & VGG-16 &ResNet-19  & ResNet-19  \\
sparsity & sparsity& CIFAR-10 & CIFAR-100 & CIFAR-10 & CIFAR-100   \\
\midrule
        & 0.9     & 90.36  &  64.52  & 89.97  &  66.09 \\
        & 0.8     & 91.02  &  65.74  & 90.21  &  67.59   \\
0.95    & 0.7     & \textbf{91.31}  &  66.57  & 90.47  &  68.30  \\
        & 0.6     & 91.11  &  66.73  & 90.56  &  \textbf{68.95}  \\
        & 0.5     & 90.94  &  \textbf{66.82}  & \textbf{90.57}  &  68.39  \\
\midrule
        & 0.9     & 89.13  &  61.92  & 88.58  &  63.25 \\
        & 0.8     & \textbf{89.62}  &  \textbf{63.5}1  & \textbf{88.61}  &  64.39   \\
0.98    & 0.7     & 89.56  &  63.21  & 88.48  &  \textbf{65.48} \\
        & 0.6     & 89.50  &  62.69  & 88.25  &  64.74 \\
        & 0.5     & 89.48  &  63.13  & 88.10  &  74.89 \\
\bottomrule
\end{tabular}
}\caption{Study on effects on different initial sparsity.}\vspace{0cm}
\label{tb:diff_init_sparsity}
\vspace{0.02cm}
\end{table}




\subsubsection{Effects of Smaller Timesteps} We compare the accuracy performance of NDSNN and LTH on a smaller timestep (i.e., $t=2$) to further validate the effectiveness of proposed method on a more efficient training approach (i.e., the smaller training timesteps, the smaller training cost in time) as shown in Fig.~\ref{fig:smaller_timestep}. It's observed that NDSNN outperforms LTH on the four experiments (i.e., VGG-16/CIFAR-10, VGG-16/CIFAR-100, ResNet-19/CIAFR-10, ResNet-19/CIAFR-100). On CIFAR-100, NDSNN has 5.55\% and 13.34\% improvements in accuracy at a sparsity of 99\% on VGG-16 and ResNet-19, respectively.





\vspace{-0.15cm}

\section{Conclusion}

In this paper, we 
propose a novel, computationally efficient, sparse training regime, \underline{N}eurogenesis
\underline{D}ynamics-inspired \underline{S}piking \underline{N}eural \underline{N}etwork training acceleration framework, NDSNN. Our proposed method trains a model from scratch using dynamic sparsity. Within our method, we create a drop-and-grow strategy which is biologically motivated by neurogenesis to promote weight reduction. Our method gives higher accuracy and is computationally less demanding than competing approaches. For example, on CIFAR-100, we can achieve an average increase in accuracy of 13.71\% over LTH for ResNet-19 across all sparsities. 
For all datasets, DNSNN has an average of 6.72\% accuracy improvement and 59.9\% training cost reduction on ResNet-19. Overall, NDSNN could shed light on energy efficient SNN training on edge devices. 


\vspace{-0.15cm}

\section*{Acknowledgement}
This work is partially supported by the National Science Foundation (NSF) under Award CCF-2011236, and Award CCF-2006748.

\vspace{-0.3cm}

\scriptsize
\bibliography{dac2023}

\end{document}